# Deep learning for smart fish farming: applications, opportunities and challenges

Xinting Yang[1,2,3], Song Zhang[1,2,3,5], Jintao Liu[1,2,3,6], Qinfeng Gao[4], Shuanglin Dong[4], Chao Zhou[1,2,3*]

1. Beijing Research Center for Information Technology in Agriculture, Beijing 100097, China
2. National Engineering Research Center for Information Technology in Agriculture, Beijing 100097, China
3. National Engineering Laboratory for Agri-product Quality Traceability, Beijing, 100097, China
4. Key Laboratory of Mariculture, Ministry of Education, Ocean University of China, Qingdao, Shandong Province, 266100, China
5. Tianjin University of Science and Technology, Tianjin 300222, China
6. Department of Computer Science, University of Almeria, Almeria, 04120, Spain

*Corresponding author: Chao Zhou    E-mail address: supperchao@hotmail.com, zhouc@nercita.org.cn

## Abstract

The rapid emergence of deep learning (DL) technology has resulted in its successful use in various fields, including aquaculture. DL creates both new opportunities and a series of challenges for information and data processing in smart fish farming. This paper focuses on applications of DL in aquaculture, including live fish identification, species classification, behavioral analysis, feeding decisions, size or biomass estimation, and water quality prediction. The technical details of DL methods applied to smart fish farming are also analyzed, including data, algorithms, and performance. The review results show that the most significant contribution of DL is its ability to automatically extract features. However, challenges still exist; DL is still in a weak artificial intelligence stage and requires large amounts of labeled data for training, which has become a bottleneck that restricts further DL applications in aquaculture. Nevertheless, DL still offers breakthroughs for addressing complex data in aquaculture. In brief, our purpose is to provide researchers and practitioners with a better understanding of the current state of the art of DL in aquaculture, which can provide strong support for implementing smart fish farming applications.

**Keywords**: Deep learning; Smart fish farming; Advanced analytics; Aquaculture;

# Contents

# 1. Introduction

In 2016, the global fishery output reached a record high of 171 million tons. Of this output, 88% is consumed directly by human beings and is essential for achieving the Food and Agriculture Organization of the United Nations (FAO)'s goal of building a world free from hunger and malnutrition (FAO, 2018). However, as the population continues to grow, the pressure on the world's fisheries will continue to increase (Merino *et al.*, 2012 ; Clavelle *et al.*, 2019).

Smart fish farming refers to a new scientific field whose objective is to optimize the efficient use of resources and promote sustainable development in aquaculture through deeply integrating the Internet of Things (IoT), big data, cloud computing, artificial intelligence and other modern information technologies. Furthermore, the real-time data collection, quantitative decision-making, intelligent control, precise investment and personalized service, have been achieved, finally forming a new fishery production mode (Figure 1).

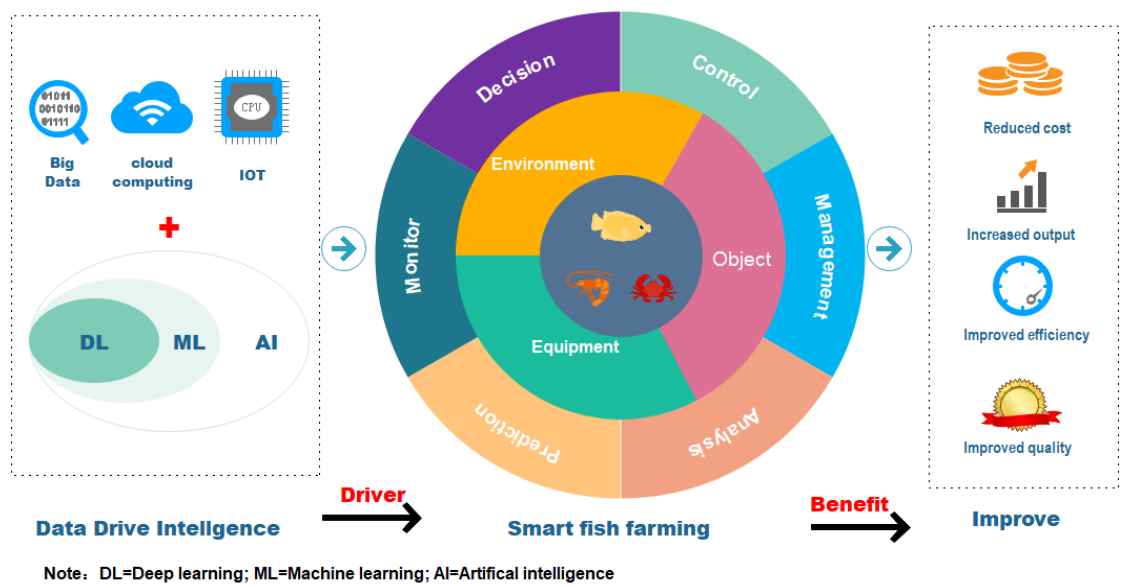

Figure 1. The role of deep learning and big data in smart fish farming

In smart fish farming, data and information are the core elements. The aggregation and advanced analytics of all or part of the data will lead to the ability to make scientifically based decisions. However, the massive amount of data in smart fish farming imposes a variety of challenges, such as multiple sources, multiple formats and complex data. Multiple sources include information regarding the equipment, the fish, the environment, the breeding process and people. The multiple formats include text, image and audio. The data complexities stem from different cultured species, modes and stages. Addressing the above high-dimensional, nonlinear and massive data is an extremely challenging task.

More attention is being paid to data and intelligence in current fish farming than ever before. As shown in Figure 1, data-driven intelligence methods, including artificial intelligence and big data, have begun to transform these data into operable information for smart fish farming (Olyaie *et al.*, 2017 ; Shahriar & McCulluch, 2014). Artificial intelligence, especially machine learning and computer vision applications, is the next frontier technology of fishery data systems (Bradley et al., 2019). Traditional machine learning methods, such as the support vector machine (SVM) (Cortes & Vapnik, 1995), artificial neural networks (ANN) (Hassoun, 1996), decision trees (Quinlan, 1986), and principal component analysis (Jolliffe, 1987), have achieved satisfactory performances in a variety of applications (Wang *et al.*, 2018). However, the conventional machine learning algorithms rely heavily on features manually designed by human engineers (Goodfellow, 2016), and it is still difficult to determine which features are most suitable for a given task (Min *et al.*, 2017).

As a breakthrough in artificial intelligence (AI), deep learning (DL) has overcome previous limitations. DL methods have demonstrated outstanding performances in many fields, such as agriculture (Yang *et al.*, 2018 ; Gouiaa & Meunier, 2017), natural language processing (Li, 2018), medicine (Gulshan *et al.*, 2016), meteorology (Mao *et al.*, 2019), bioinformatics (Min *et al.*, 2017), and security monitoring (Dhiman & Vishwakarma, 2019). DL belongs to the field of machine learning but improves data processing by extracting highly nonlinear and complex features via sequences of multiple layers automatically rather than requiring handcrafted optimal feature representations for a particular type of data based on domain knowledge (LeCun *et al.*, 2015 ; Goodfellow, 2016). With its automatic feature learning and high-volume modeling capabilities, DL provides advanced analytical tools for revealing, quantifying and understanding the enormous amounts of information in big data to support smart fish farming (Liu *et al.*, 2019). DL techniques can be used to solve the problems of limited intelligence and poor performance in the analysis of massive, multisource and heterogeneous big data in aquaculture. By combining the IoT, cloud computing and other technologies, it is possible to achieve intelligent data processing and analysis, intelligent optimization and decision-making control functions in smart fish farming.

This paper provides a comprehensive review of DL and its applications in smart fish farming. First, the various DL applications related to aquaculture are outlined to highlight the latest advances in relevant areas, and the technical details are briefly introduced. Then, the challenges and future trends of DL in smart fish farming are discussed. The remainder of this paper is organized as follows: After the Introduction, Section 2 introduces basic background knowledge such as DL terminology, definitions, and the most popular learning models and algorithms. Section 3 describes the main applications of DL in aquaculture, and Section 4 provides technical details. Section 5 discusses the advantages, disadvantages and future trends of DL in smart fish farming, and Section 6 concludes the paper.

## 2. Concepts of deep learning

### 2.1 Terms and definitions of deep learning

Machine learning (ML), which emerged together with big data and high-performance computing, has created new opportunities to unravel, quantify, and understand data-intensive processes. ML is defined as a scientific field that seeks to give machines the ability to learn without being strictly programmed (Samuel, 1959 ; Liakos *et al.*, 2018). Deep learning is a branch of machine learning and is type of representation learning algorithm based on an artificial neural network (Deng & Yu, 2014). Specifically, DL is a type of machine learning that can be used for many (but not all) AI tasks (Goodfellow, 2016 ; Saufi *et al.*, 2019).

DL enables computers to build complex concepts from simpler concepts, thus solving the core problem of representation learning (Bronstein *et al.*, 2017 ; LeCun *et al.*, 2015). Figure 2 shows an example of how a DL system might represent the concept of a fish in an image by combining simpler concepts. It is difficult for computers to directly understand the meaning contained in raw sensory input data, such as an image represented as a set of pixels. The functions that map a set of pixels to an object are highly complex. It seems impossible to learn or evaluate such a mapping through direct programming. To solve this problem, DL decomposes this complex mapping into a nested series of simpler mappings. For example, an image is input in the visible layer, followed by a series of hidden layers that extract increasingly abstract features from the image. Given a pixel, by comparing the brightness of adjacent pixels, the first layer could easily identify whether this pixel represents an edge. Then, the second hidden layer searches for sets of edges that can be recognized as angles and extended contours. The third hidden layer can then find a specific set of contours and corners that represent an entire portion of a particular object. Finally, the various objects existing in the image can be identified (Goodfellow, 2016 ; Zeiler & Fergus, 2014).

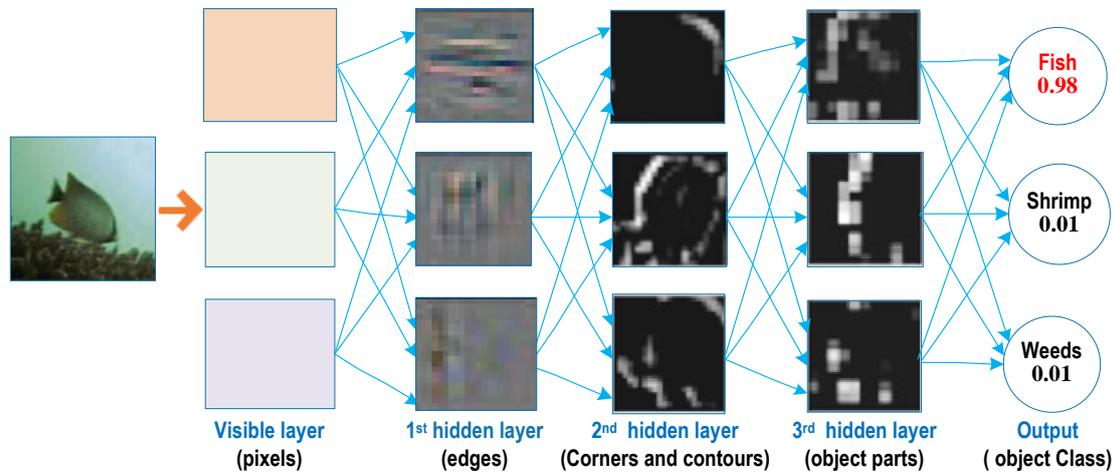

Figure 2. An example of a DL model

## 2.2 Learning tasks and models

In general, a DL method involves a learning process whose purpose is to gain "experience" from samples to support task execution. DL methods can be divided into two categories: supervised learning and unsupervised learning (Goodfellow, 2016). In supervised learning, data are presented as labeled samples consisting of inputs and corresponding outputs. The goal is to construct mapping rules from the input to output. The convolutional neural network (CNN) and the recurrent neural network (RNN) are two typical popular model architectures. Inspired by the human visual nervous system, CNNs excel at image processing (Ravì *et al.*, 2016 ; Saufi *et al.*, 2019 ; Litjens *et al.*, 2017), while an RNN can process sequential data effectively. In unsupervised learning, the data are not labeled; instead the model seeks previously undetected patterns in a dataset with no pre-existing labels and with minimal human supervision (Geoffrey E Hinton, 1999). The generative adversarial network (GAN) is one of the most promising unsupervised learning approaches. A GAN can produce good output through mutual game learning of two (at least) modules in the framework: a generative model and a discriminative model. Many modified or improved models have been derived based on these original DL models, such as the region convolutional neural network (R-CNN) and long short-term memory (LSTM) models.

Figure 3 shows a comparison of traditional machine learning and DL. In DL, feature learning and model construction are integrated into a single model via end-to-end optimization. In traditional machine learning, feature extraction and model construction are performed separately, and each module is constructed in a step-by-step manner.

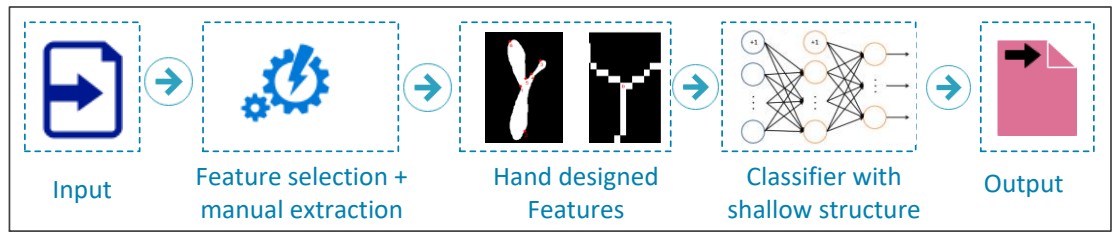

(a) Machine learning

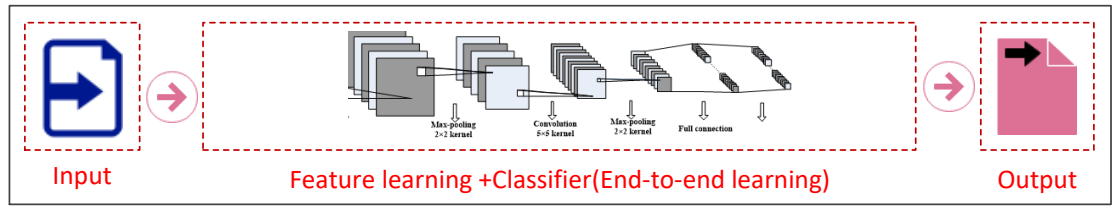

(b) Deep learning

Figure 3. Comparison of DL and machine learning

Compared with the shallow structure of traditional machine learning, the deep hierarchical structure used in DL makes it easier to model nonlinear relationships through combinations of functions (Liakos *et al.*, 2018 ; Wang *et al.*, 2018). The advantages of DL are especially obvious when the amount of data to be processed is large. More specifically, the hierarchical learning and extraction of different levels of complex data abstractions in DL provides a certain degree of simplification for big data analytics tasks, especially when analyzing massive volumes of data, performing data tagging, information retrieval, or conducting discriminative tasks such as classification and prediction (Najafabadi *et al.*, 2015). Hierarchical architecture learning systems have achieved superior performances in several engineering applications (Poggio & Smale, 2003 ; Mhaskar & Poggio, 2016).

The overall structure, process and principles of applying deep learning to fishery management is depicted in Figure 4. After the data are collected and transmitted, deep learning performs inductive analysis, learns the experience or knowledge from the samples, and finally formulates rules to guide management decisions.

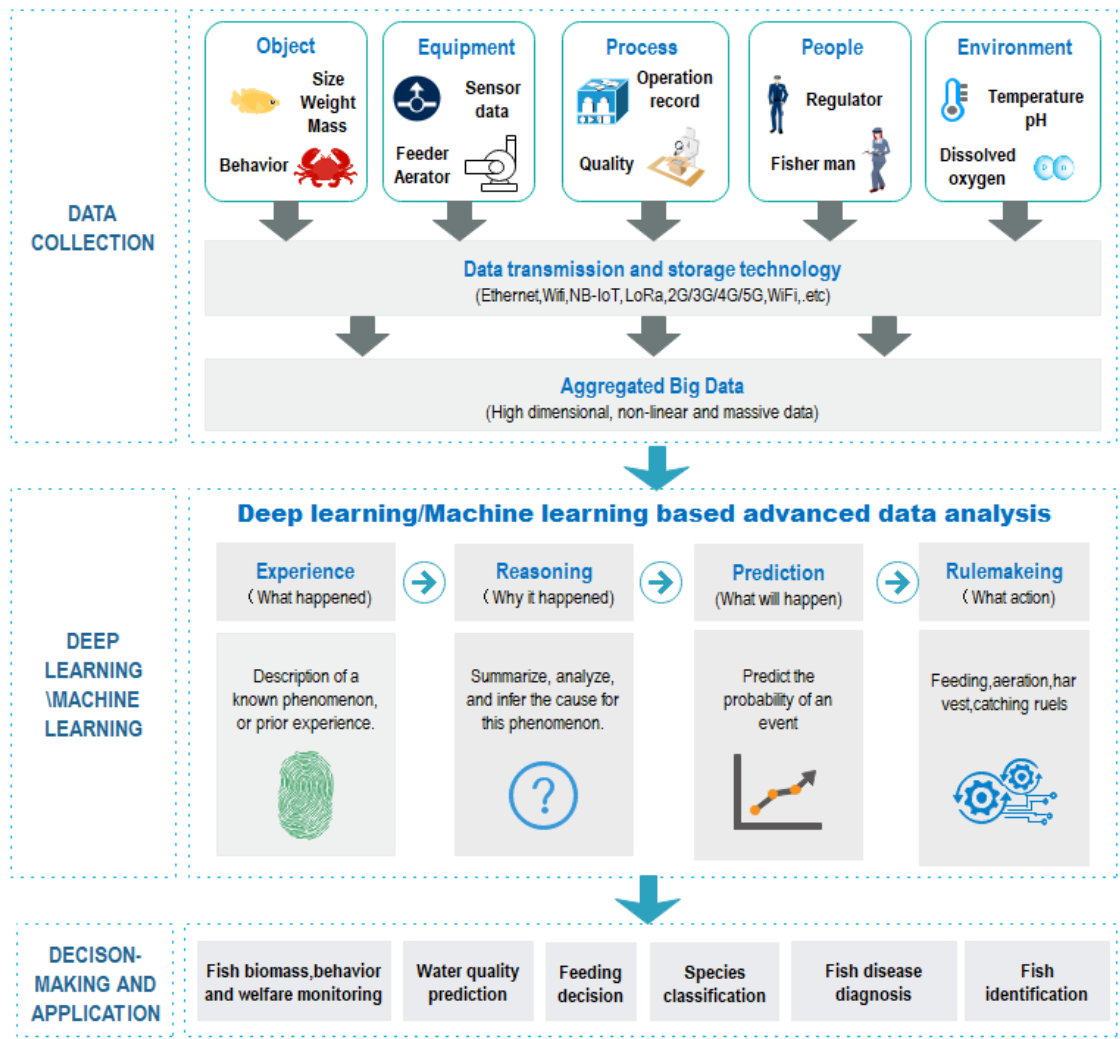

Figure 4. Deep-learning-enabled advanced analytics for smart fish farming

However, when applying deep learning, the most serious issue is that of hallucination. Another failure mode of neural networks is overlearning or overfitting. In addition, neural networks can be tricked into producing completely different outputs after imperceptible perturbations are applied to their inputs (Belthangady & Royer, 2019 ; Moosavi-Dezfooli *et al.*, 2016).

## 3. Applications of deep learning in smart fish farming

This review discussed 41 papers related to DL and smart fish farming. The relevant applications can be divided into 6 categories: live fish identification, species classification, behavioral analysis, feeding decisions, size or biomass estimation, and water quality prediction. Figure 5 shows the number of papers related to each application. The most popular fields are live fish identification and species classification. Notably, all these papers were published in 2016 or later, including 3 in 2016, 3 in 2017, 12 in 2018, 15 in 2019, and 8 in 2020 (through May 2020), indicating that DL has developed rapidly since 2016. In addition to water quality prediction and sound recognition, most papers involve image processing. Moreover, while most of the papers are focused on fish, a few works consider lobsters or other aquatic animals.

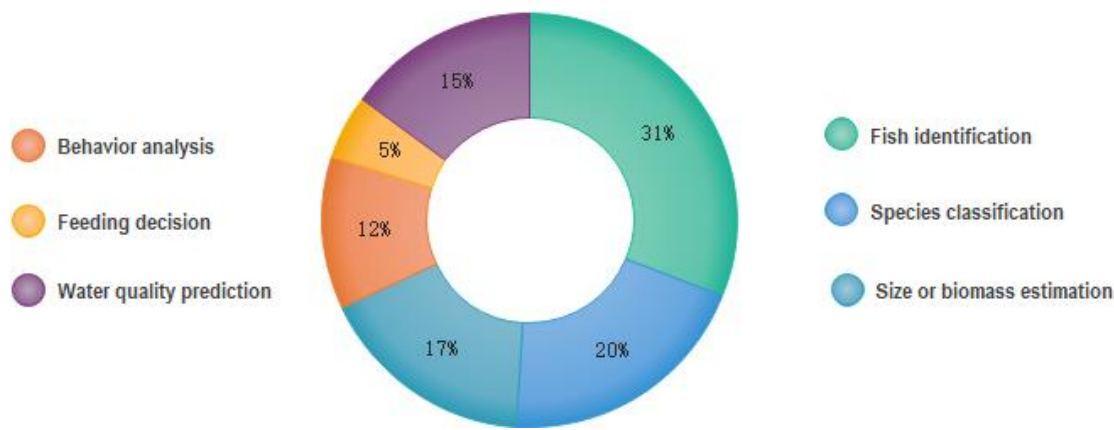

Figure 5. Numbers of papers addressing different application scenarios

## 3.1 Live fish identification

Accurate and automatic live fish identification can provide data support for subsequent production management; thus, fish identification is an important factor in the development of intelligent breeding management equipment or systems. Machine vision has the advantages of enabling long-term, nondestructive, noncontact observation at low cost (Zhou *et al.*, 2018b ; Hartill *et al.*, 2020). However, the scenes encountered in aquaculture present numerous challenges for image and video analysis. First, the image quality is easily affected by light, noise, and water turbidity, resulting in relatively low resolution and contrast (Zhou *et al.*, 2017a). Second, because fish swim freely and are uncontrolled targets, their behavior may cause distortions, deformations, occlusion, overlapping and other disadvantageous phenomena (Zhou *et al.*, 2017b). Most current image analysis methods are adversely affected by these difficulties (Qin *et al.*, 2016 ; Sun *et al.*, 2018).

While many studies have been conducted to investigate the above issues, most emphasized the extraction of conventional low-level features, which usually involve small details in an image such as feature points, colors, textures, contours, and shapes of interest (White *et al.*, 2006 ; Yao & Odobez, 2007). In practical applications, the effects of methods based on such features is often unsatisfactory. DL involves multilevel data representations, from low to high levels, in which high-level features are built on the low-level features and carry rich semantic information that can be used to recognize and detect targets or objects in the image. Generally, both types of features are used in convolutional neural networks: the first few layers of learn the low-level features, and the last few layers learn the high-level features. This approach has the potential to solve the problems listed above (Sun *et al.*, 2018 ; Zheng *et al.*, 2017).

Table 1 shows the details of live fish identification using DL. CNNs can be used to extract features from fish or shrimp images (Hu *et al.*, 2020). By training on a public dataset with real images, compared with SVM and Softmax, the CNN model identification accuracy improved by 15% and 10%, respectively, making automatic recognition more accurate (Qin *et al.*, 2016). Although the aforementioned CNN architecture shows good performance, a CNN detects features using sliding window, which can waste resources. To overcome the above challenges, a region-based CNN (R-CNN) can be used to detect freely moving fish in an unconstrained underwater environment. An R-CNN judges object locations by extracting multiple region proposals and then applying a CNN to only the best candidate regions, which improves model efficiency (Girshick *et al.*, 2014). The candidate fish-containing regions can be generated via both fish motion information and from the raw image (Salman *et al.*, 2019). The advantage of R-CNN is that it improves the accuracy by at least 16% over a Gaussian mixture model (GMM) on the FCS dataset.

Because classical CNNs are trained through supervised learning, their recognition capability depends primarily on the quality of the training samples and their annotations (LeCun *et al.*, 2015). A semisupervised DL model can learn not only from labeled samples but also from unlabeled data. Thus, a GAN can somewhat alleviate the challenges posed by a lack of labeled training data in practical applications (Zhao *et al.*, 2018b). Using a synthetic dataset, Mahmood *et al.* (2019) trained the You Only Look Once (YOLO) v3 object detector to detect lobsters in challenging underwater images, thus addressing a problem involving complex body shapes, partially accessible local environments, and limited training data. In some cases, even when insufficient training data is available, a transfer framework can be used to effectively learn the characteristics of underwater targets with the help of data enhancement. Data enhancement improves the data quality by adjusting the contrast, entropy, and other factors in images or it expands the number of samples via operations such as flipping, translation or rotation. The increased variety and number of samples allow models to achieve higher accuracy (Sun *et al.*, 2018).

To meet the needs of some embedded systems, such as underwater drones, real-time performance by DL models are the key to their practicability. It has been experimentally shown that using an unmanned aerial vehicle (UAV)-type system to observe objects on the sea surface, a CNN can effectively recognize a swarm of jellyfish, and can achieve reasonable performance levels (80% accuracy) for real-world applications (Kim *et al.*, 2016). After DL model training is complete, such models can show excellent speed for live fish identification purposes. For example, one model required only 6 s to identify 115 images (Meng *et al.*, 2018); the average time to detect lionfish in each frame was only 0.097 s (Naddaf-Sh *et al.*, 2018). Therefore, under the premise of reasonable accuracy, a DL model's recognition speed can satisfy real-time requirements (Villon *et al.*, 2018). Hence, DL can be effectively applied to identify fish while meeting the rapid response and real-time requirements of embedded systems.

For identifying live fish, DL is mainly used to solve the problem of whether a given object is a fish (Ahmad et al., 2016). In this era, where large amounts of visual data can be collected easily, DL can be a practical machine vision solution. Therefore, it is worth studying the performance levels that can be achieved by combining DL and machine vision to explore fast and accurate methods. The main disadvantage of DL is that it requires a large amount of labeled training data, and obtaining and annotating sufficiently large numbers of images is time-consuming and laborious. Moreover, the recognition effect depends on the quality of the training samples and annotations.

Table 1 Live fish identification

| | | Model | Frame work | Data | Preprocessing augmentation | Transfer learning | Evaluation index | Results | Comparisons with other methods |
|---|---|---|---|---|---|---|---|---|---|
| 1 | Qin *et al.* (2016) | CNN | Caffe | Fish4Knowledge (F4K) dataset | Resize Rotation | N | Accuracy | Accuracy: 98.64% | LDA+SVM: 80.14%; Raw-pixel Softmax: 87.56%; VLFeat Dense-SIFT: 93.56% |
| 2 | Zhao *et al.* (2018b) | DCGAN | Tensor Flow | F4K dataset, Croatian fish dataset | Image segmentation and enhancement | N | Accuracy | Accuracy: 83.07%. | Accuracy: CNN: 72.09%, GAN: 75.35% |
| 3 | Sun *et al.* (2018) | CNN | Caffe | F4K dataset | Horizontal mirroring, crop | Y | Precision(P), recall(R) | P: 99.68%; R: 99.45% | P: Gabor: 58.55%; Dsift-Fisher: 83.37%; LDA: 80.14%; DeepFish: 90.10%; RGB-Alex-SVM: 99.68% |
| 4 | Meng *et al.* (2018) | CNN | NA | 4 kinds of fish and 100 images of every kind selected from Google. | Blur, rotation | N | Accuracy, speed | Accuracy: 87%, Speed: 115 f/6s. | Accuracy: AlexNet:87%; GoogLeNet: 85%, LeNet: 67% |
| 5 | Naddaf-Sh *et al.* (2018) | CNN | NA | Videos collected with an ROV camera; 1,500 images were gathered from online resources such as ImageNet, Google and YouTube | Resize | N | True Positive, False Positive, speed | TPR:93%; FPR:4%; Speed: 0.097s/f | NA |
| 6 | Villon *et al.* (2018) | CNN | Caffe | 5 frames per second were extracted, leading to a database of 450,000 frames. | NA | N | Accuracy, Speed | Accuracy：94.9%, each identification took 0.06 s. | Average success rate: Humans:89.3% |
| 7 | Kim *et al.* (2016) | CNN | NA | The image set was obtained using a UAV. | NA | N | TPR, FPR | TPR: 0.80 FPR: 0.04 | NA |
| 8 | Salman | CNN | Tensor | F4K dataset, LCF-15 dataset | NA | Y | Accuracy | F4K: 87.44%; | GMM：71.01%; |
| | *et al.* (2019) | | Flow | | | | | LCF-15: 80.02% | Optical flow: 56.13%; R-CNN：64.99% |
| 9 | Labao and Naval (2019) | R-CNN | NA | 10 underwater video sequences for a total of 300 training frames | NA | N | Precision, Recall, F-Score | Accuracy increased by correction mechanism | NA |
| 10 | Mahmood *et al.* (2019) | Yolo | Darknet | The dataset was generated and synthesized by using the ImageNet dataset | NA | N | Mean average precision | The synthetic data can achieve higher performance than the baseline. | NA |
| 11 | Guo *et al.* (2019) | DRN | PyTorch | The dataset was composed of 908 negative and 907 positive samples | resize | N | accuracy | higher than 82% | |
| 12 | Hu *et al.* (2020) | CNN | Keras | 16,138 samples were collected from Google, and self-shot videos. | Resized, grayscale | N | Accuracy | 95.48% | NA |
| 13 | Cao *et al.* (2020) | CNN | Tensor Flow | The video was acquired from a crab-breeding operation in Jiangsu province | image denoising and enhancement | N | precision (AP) | AP: 99.01%; F1: 98.74% | AP：YOLOV3：93.73%；Faster RCNN：99.05%；F1：YOLOV3：92.47%；Faster RCNN：98.56%；HOG + SVM：73.18%； |

## 3.2 Species classification

Fish are diverse, with more than 33,000 species (Oosting *et al.*, 2019). In aquaculture, species classification is helpful for yield prediction, production management, and ecosystem monitoring (Alcaraz *et al.*, 2015； dos Santos & Gonçalves, 2019). Fish species can usually be distinguished by visual features such as size, shape, and color (dos Santos & Gonçalves, 2019； Hu *et al.*, 2012). However, due to changes in light intensity and fish motion as well as similarities in the shapes and patterns among different species, accurate fish species classification is challenging.

DL models can learn unique visual characteristics of species that are not sensitive to environmental changes and variations. Table 2 shows some details when using DL. Taking a given underwater video as an example (Figure 6), an object detection module first generates a series of patch proposals for each frame $F$. Each patch is then used as an input to the classifier, and a label distribution vector is obtained. The tags with the highest probability are regarded as the tags of these patches (Sun *et al.*, 2018).

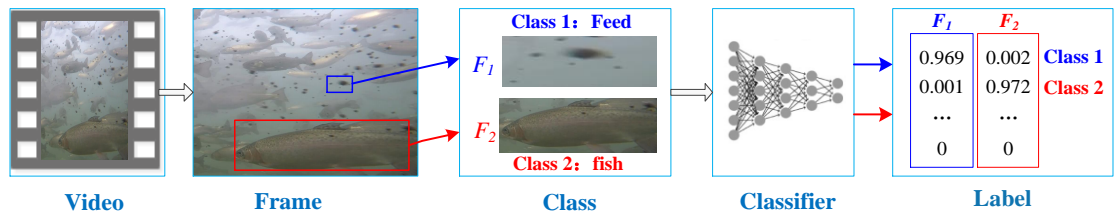

Figure 6. An illustration of the fish classification process

A DL model can better distinguish differences in characteristics, categories, and the environment, which can be used to extract the features of target fish from an image collected in an unconstrained underwater environment. Fish species can be classified to identify several basic morphological features (i.e., the head region, body shape, and scales) (Rauf *et al.*, 2019). Most of the DL models show better results compared with the traditional approaches, reaching classification accuracies above 90% on the LifeCLEF 14 and LifeCLEF 15 benchmark fish datasets (Ahmad *et al.*, 2016). To avoid the need for large amounts of annotated data, general deep structures must be fine-tuned to improve the effectiveness with which they can identify the pertinent information in the feature space of interest. Accordingly, various DL models for identifying fish species have been developed using a pretrained approach called transfer learning (Siddiqui *et al.*, 2017； Lu *et al.*, 2019； Allken *et al.*, 2019). By fine-tuning pretrained models to perform fish classification using small-scale datasets, these approaches enable the network to learn the features of a target dataset accurately and comprehensively (Qiu *et al.*, 2018), and achieved sufficiently high accuracy to serve as economical and effective alternatives to manual classification.

In addition to visual characteristics, different species of grouper produce different sound frequencies that can be used to distinguish these species. For example, CNN and LSTM models were used to classify sounds produced by four species of grouper; their resulting classification accuracy was significantly better than the previous weighted mel-frequency cepstral coefficients (WMFCCs) method (Ibrahim *et al.*, 2018).

Nevertheless, due to the influence of various interferences and the small sets of available samples, the accuracy of same-species classification still has considerable room to improve. Most current fish classification methods are designed to distinguish fish with significant differences in body size or shape; thus, the classification of similar fish and fish of the same species is still challenging (dos Santos & Gonçalves, 2019).

Table 2 Species classification

| | | Model | Framework | Data | Preprocessing augmentation | Transfer learning | Evaluation index | Results | Comparisons with other methods |
|---|---|---|---|---|---|---|---|---|---|
| 1 | Siddiqui et al. (2017) | CNN | MatConvNet | Videos were collected from several baited remote underwater video sampling programs during 2011–2013. | Resized | Y | Accuracy | 94.3% | SRC: 65.42%; CNN: 87.46% |
| 2 | Ahmad et al. (2016) | CNN | NA | LifeCLEF14 and LifeCLEF15 dataset | Resized and converted to grayscale. | N | Precision, and Recall | AC>90%; each fish image takes approximately 1 ms for classification. | SVM, KNN, SRC, PCA-SVM, PCA-KNN, CNNSVM, CNN-KNN |
| 3 | Ibrahim et al. (2018) | LSTM and CNN | NA | The dataset contains 60,000 files, and the audio duration of each file is 20 s at a sampling rate of 10 kHz. | NA | N | Accuracy | 90% | WMFCC<90% |
| 4 | Qiu et al. (2018) | CNN | NA | ImageNet dataset, F4K dataset, a small-scale fine-grained dataset (*i.e.*, Croatian or QUT fish dataset). | Super resolution, Flip and rotation | Y | Accuracy | 83.92% | B-CNNs: 83.52%; B-CNNs+SE BLOCKS: 83.78% |
| 5 | Allken et al. (2019) | CNN | TensorFlow | ImageNet classification dataset and the images collected by the Deep Vision system; a total of 1,216,914 stereo image pairs from 63 h 19 min of | Resized; Rotation, translation, shearing, flipping, and zooming | Y | Accuracy | 94% | NA |

| | | | | | | | | | | |
|---|---|---|---|---|---|---|---|---|---|---|
| | | | | | data collection. | | | | | |
| 6 | Rauf et al. (2019) | CNN | NA | Fish-Pak | Resize; Image background transparent | Y | Accuracy, Precision, Recall, F1-Score | The proposed method achieves state of the art performance and outperforms existing methods | VGG-16, one block VGG, two block VGG, three block VGG, LeNet-5, AlexNet, GoogleNet, and ResNet-50 | |
| 7 | Lu et al. (2019) | CNN | NA | A total of 16,517 fish catching images were provided by Fishery Agency, Council of Agriculture (Taiwan) | Resize; Horizontal flipping, vertical flipping, width shifting, height shift, rotation, shearing, zoom-in, and zoom-out | Y | Accuracy | > 96.24%. | NA | |
| 8 | Jalal et al. (2020) | YOLO, CNN | Tensor Flow | LCF15 datasheet and UWA datasheet | NA | N | Accuracy | LCF15: 91.64%' UWA: 79.8% | | |

## 3.3 Behavioral analysis

Fish are sensitive to environmental changes, and they exhibit a series of responses to changes environmental factors through behavioral changes (Saberioon *et al.*, 2017； Mahesh *et al.*, 2008). In addition, behavior serves as an effective reference indicator for fish welfare and harvesting (Zion, 2012). Relevant behavior monitoring, especially for unusual behaviors, can provide a nondestructive understanding and an early warning of fish status (Rillahan *et al.*, 2011). Real-time monitoring of fish behavior is essential in understanding their status and to facilitate capturing and feeding decisions (Papadakis *et al.*, 2012).

Fish display behavior through a series of actions that have a certain continuity and time correlations. Methods of identifying an action from a single image will lose relevance for images acquired before and after the action. Therefore, it is desirable to use time-series information related to the prior and subsequent frames in a video to capture action relevance. DL methods have shown strong ability to recognize visual patterns (Wang *et al.*, 2017). Table 3 shows the details of the behavioral analysis using DL. In particular, due to their powerful modeling capabilities for sequential data, RNNs have the potential to address the above problem effectively (Schmidhuber, 2015). Zhao *et al.* (2018a) proposed a novel method based on a modified motion influence map and an RNN to systematically detect, localize and recognize unusual local behaviors of a fish school in intensive aquaculture.

Tracking individuals in a fish school is a challenging task that involves complex nonrigid deformations, similar appearances, and frequent occlusions. Fish heads have relatively fixed shapes and colors that can be used to track individual fish (Butail & Paley, 2011； Wang *et al.*, 2012). Thus, data associations can be achieved across frames, and as a result, behavior trajectory tracking can be implemented without being affected by frequent occlusions (Wang *et al.*, 2017). In addition, data enhancement and iterative training methods can be used to optimize the accuracy of classification tasks for identifying behaviors that cannot be distinguished by the human eye (Xu & Cheng, 2017). Finally, idTracker and further developments in identification algorithms for unmarked animals have been successful for 2~15 individuals in small groups (Pérez-Escudero *et al.*, 2014). An improved algorithm, called Idtracker.ai has also been proposed. Using two different CNNs, Idtracker.ai can track all the individuals in both small and large groups (up to 100 individuals) with a recognition accuracy that typically exceeds 99.9% (Romero-Ferrero *et al.*, 2019).

When using deep learning to classify fish behavior, crossing, overlapping and blocking caused by free-swimming fish (Zhao *et al.*, 2018a ; Romero-Ferrero *et al.*, 2019) and low-quality environmental images (Zhou et al., 2019) form the main challenges to behavior analysis; thus, these problems need to be solved in the future.

Table 3 Behavior analysis

| | Field | Model | Framework | Data | Preprocessing augmentation | Transfer learning | Evaluation index | Results | Comparisons with other methods |
|---|---|---|---|---|---|---|---|---|---|
| 1 | Xu and Cheng (2017) | CNN | MatCovNet | The head feature maps stored in the segment in the trajectory along with the trajectory ID form the initial training dataset. | Shifting, horizontal and vertical rotation | N | Precision, Recall, F1-measure, MT, ML, Fragments, ID Switch | The proposed method performs significantly well on all metrics. | NA |
| 2 | Zhao et al. (2018a) | RNN | Tensor Flow | The behavior dataset was made manually following All Occurrences Sampling. | NA | N | Accuracy | detection, localization and recognition: 98.91%, 91.67% and 89.89% | Accuracy of OMIM and OMIM less than 82.45% |
| 3 | Wang et al. (2017) | CNN | MatCovNet | Randomly selected 300 frames from each of the 5 datasets and manually annotated the head point in each frame. | rotated | N | IR, Miss ratio, Error ratio, Precision, recall, MT, ML, Frag, IDS | The proposed method outperforms two state-of-the-art fish tracking methods in terms of 7 performance metrics | idTracker |
| 4 | Romero-Ferrero et al. (2019) | CNN | NA | 184 juvenile zebrafish, the dataset comprised 3,312,000 uncompressed, grayscale, labeled images. | extracts 'blobs', and then oriented | Y | Accuracy | 99.95% | NA |
| 5 | Li et al. (2020) | CNN | Tensor Flow | The image was collected from a glass aquarium | Cut and synthesis | N | Accuracy, precision and recall, | Accuracy: 99.93%, precision: 100%, recall: 99.86% | |

## 3.4 Size or biomass estimation

It is essential to continuously observe fish parameters such as abundance, quantity, size, and weight when managing a fish farm (França Albuquerque *et al.*, 2019). Quantitative estimation of fish biomass forms the basis of scientific fishery management and conservation strategies for sustainable fish production (Zion, 2012 ; Li *et al.*, 2019 ; Saberioon & Císař, 2018 ; Lorenzen *et al.*, 2016 ; Melnychuk *et al.*, 2017). However, it is difficult to estimate fish biomass without human intervention because fish are sensitive and move freely within an environment where visibility, lighting and stability are typically uncontrollable (Li *et al.*, 2019).

Recent applications of DL to fishery science offer promising opportunities for massive sampling in smart fish farming. Machine vision combined with DL can enable more accurate estimation of fish morphological characteristics such as length, width, weight, and area. Most reported applications have been either semisupervised or supervised (Marini *et al.*, 2018 ; Díaz-Gil *et al.*, 2017). For example, the Mask R-CNN architecture was used to estimate the size of saithe (*Pollachius virens*), blue whiting (*Micromesistius poutassou*), redfish (*Sebastes spp.*), Atlantic mackerel (*Scomber scombrus*), velvet belly lanternshark (*Etmopterus spinax*), Norway pout (*Trisopterus esmarkii*), Atlantic herring (*Clupea harengus*) (Garcia *et al.*, 2019) and European hake (Álvarez-Ellacuría *et al.*, 2019). Another method for indirectly estimating fish size is to first detect the head and tail of fish with a DL model and then calculate the length of fish on that basis. Although this approach increases the workload, it is suitable for more complex images (Tseng *et al.*, 2020). The structural characteristics and computational capabilities of DL models can be fully exploited (Hu *et al.*, 2014) to achieve superior performances compared with other models. In addition, DL-based methods can eliminate the influence of fish overlap during length estimation.

The number of fish shoals can also provide valuable input for the development of intelligent systems. DL has shown comprehensive advantages in animal computing. To achieve automatic counting of fish groups under high density and frequent occlusion characteristics, a fish distribution map can be constructed using DL; then, the fish distribution, density and quantity can be obtained. These values can indirectly reflect fish conditions such as starvation, abnormalities and other states, thereby providing an important reference for feeding or harvest decisions (Zhang *et al.*, 2020).

The age structure of a fish school is another important input to fishery assessment models. The current method for determining fish school age structure relies on manual assessments of otolith age, which is a labor-intensive and expertise-dependent process. Using a DL approach, target recognition can instead be performed by using a pretrained CNN to estimate fish ages from otolith images. The accuracy is equivalent to that achieved by human experts and considerably faster (Moen *et al.*, 2018).

Optical imaging and sonar are often used to monitor fish biomass. A DL algorithm can be applied to automatically learn the conversion relationship between sonar images and optical images, thus allowing a "daytime" image to be generated from a sonar image and a corresponding night vision camera image. This approach can be effectively used to count fish, among other applications (Terayama *et al.*, 2019).

Table 4 Size or biomass estimation

| | | Model | Framework | Data | Preprocessing and augmentation | Transfer learning | Evaluation index | Results | Comparisons with other methods |
|---|---|---|---|---|---|---|---|---|---|
| 1 | Levy et al. (2018) | CNN | Keras | ILSVRC12 (Imagenet) dataset | NA | Y | Accuracy | The method is robust and can handle different types of data, and copes well with the unique challenges of marine images. | YOLO network topology |
| 2 | Terayama et al. (2019) | GAN | NA | 1,334 camera and sonar image pairs from 10 min of data at acquired at 3 fps | Resized; normalized; flipped | N | NA | The proposed model successfully generates realistic daytime images from sonar and night camera images. | NA |
| 3 | Moen et al. (2018) | CNN | TensorFlow | The dataset comprises 4,109 images of otolith pairs and 657 images of single otoliths, totaling 8,875 otoliths. | Rotated and normalization | N | MSE, MCV | Mean CV: 8.89%: lowest MSE value: 2.65 | Comparing accuracy to human experts, mean CV of 8.89% |
| 4 | Álvarez-Ellacuría et al. (2019) | R-CNN | NA | COCO dataset; Photos were obtained with a single webcam, resolution: 1,280×760. | NA | Y | Root-mean-square deviation | 1.9 cm | NA |
| 5 | Zhang et al. (2020) | CNN | Keras | Data were collected from the "Deep Blue No. 1" net cage. The resolution is 1,920×1,080 and frame | Resized and enhanced; Gaussian noise and salt-and- | N | Accuracy | Accuracy: 95.06% | CNN: 89.61% MCNN: 91.18% |

| | | | | rate is 60 fps. | pepper noise were added | | | | |
|---|---|---|---|---|---|---|---|---|---|
| 6 | Tseng et al. (2020) | CNN | Keras | 9,000 fish images were provided by Fisheries Agency, Council of Agriculture (Taiwan). Another dataset of 154 fish images was acquired at Nan-Fang-Ao fishing harbor (Yilan, Taiwan). | Resized; Rotation, horizontal and vertical shifting, horizontal and vertical flipping, and scaling | N | Accuracy | Accuracy: 98.78% | NA |
| 7 | Fernandes et al. (2020) | CNN | | The dataset with 1,653 fish images was acquired using a Sony DSCWX220 digital camera, | NA | | $R^2$ | $R^2$: BW: 0.96, CW: 0.95 | NA |

## 3.5 Feeding decision-making

In intensive aquaculture, the feeding level of fish directly determines the production efficiency and breeding cost (Chen *et al.*, 2019). In actual production, the feed cost for some varieties of fish accounts for more than 60% of the total cost (de Verdal *et al.*, 2017 ; Føre *et al.*, 2016 ; Wu *et al.*, 2015). Thus, unreasonable feeding will reduce production efficiency, while insufficient feeding will affect fish growth. Excessive feeding also reduces the feed conversion efficiency, and the residual bait will pollute the environment (Zhou *et al.*, 2018a). Therefore, large economic benefits can be gained by optimizing the feeding process (Zhou *et al.*, 2018c). However, many factors affect fish feeding, including physiological, nutritional, environmental and husbandry factors; consequently it is difficult to detect the real needs of fish (Sun *et al.*, 2016).

Traditionally, feeding decisions depend primarily on experience and simple timing controls (Liu *et al.*, 2014b). At present, most research on making feeding decisions using DL has focused mostly on image analysis. By using machine vision, an improved feeding strategy can be developed in accordance with fish behavior. Such a system can terminate the feeding process at more appropriate times, thereby reducing unnecessary labor and improving fish welfare (Zhou *et al.*, 2018a). The feeding intensity of fish can also be roughly graded and used to guide feeding. A combination of CNN and machine vison has proved to be an effective way to assess fish feeding intensity characteristics (Zhou *et al.*, 2019); the trained model accuracy was superior to that of two manually extracted feature indicators: flocking index of fish feeding behavior (FIFFB) and snatch intensity of fish feeding behavior (SIFFB) (Zhou *et al.*, 2017b ; Chen *et al.*, 2017). This method can be used to detect and evaluate fish appetite to guide production practices. Due to recent advances in CNNs, it would be interesting to consider the use of newer neural network frameworks for both spatial and motion feature extraction. When combined with time-series information, such models may enable better feeding decisions. Based on this idea, Måløy *et al.* (2019) considered both temporal and spatial flow by combining a three-dimensional CNN (3D-CNN) and an RNN to form a new dual deep neural network. The 3D-CNN and RNN were used to capture spatial and temporal sequence information, respectively, thereby achieving recognition of both feeding and nonfeeding behaviors. A comparison showed that the recognition results achieved with this dual-flow structure were better than those of either individual CNN or RNN models.

The studies discussed above focused primarily on images. However, many factors affect fish feeding (Sun *et* al., 2016); consequently, considering only images is insufficient. In the future, additional data, such as environmental measurements and fish physiological data, will need to be incorporated to achieve more reasonable feeding decisions.

Table 5 Feeding decisions

| | Model | Framework | Data | Preprocessing augmentation | Transfer learning | Results | Performance comparison |
|---|---|---|---|---|---|---|---|
| 1 | Måløy et al. (2019) | RNN | Tensor Flow | 76 videos taken at a resolution of 224×224 pixels with RGB color channels and at 24 f/sec. | NA | N | Accuracy : 80% | NA |
| 2 | Zhou et al. (2019) | CNN | NA | Image was collected from a laboratory at 1 f/sec. | RST | N | Accuracy :90%; | SVM: 73.75%; BPNN: 81.25%; FIFFB: 86.25%; SIFFB: 83.75% |

## 3.6 Water quality prediction

It is essential to be able to predict changes in water quality parameters to identify abnormal phenomena, prevent disease, and reduce the corresponding risks to fish (Hu *et al.*, 2015). In real-world aquaculture, the water environment is characterized by many parameters that affect each other, causing considerable inconvenience in the prediction process (Liu *et al.*, 2014a). The traditional machine-learning-based prediction models lack robustness when applied to big data, resulting in a general lack of long-term modeling capability and generalizability, and they cannot fully reflect the essential characteristics of the data (Liu *et al.*, 2019； Ta & Wei, 2018). In contrast, DL offers good capabilities in terms of nonlinear approximation, self-learning, and generalization. In recent years, prediction methods based on DL have been widely used (Roux & Bengio, 2008).

Dissolved oxygen is one of the most important parameters and is important in intelligent management and control in smart fish farming (Rahman *et al.*, 2019). Due to the time lag between the implementation of control measures for dissolved oxygen and their regulation effects, it is necessary to predict future changes in dissolved oxygen to maintain a stable water quality (Ta & Wei, 2018). DL-based models such as a CNN or a deep belief network (DBN) can extract the relationships between quantitative water characteristics and water quality variables (Lin *et al.*, 2018). Such models have been used to predict water quality parameters for the intensive culturing of fish or shrimp. The results show that the accuracy and stability of such models are sufficient to meet actual production needs (Ta & Wei, 2018).

However, most current methods have achieved good results only for short-term water quality predictions. In recent years, scholars have paid increasing attention to longer-term predictions. The key to long-term prediction is to extract the spatiotemporal relationships between water quality and external factors. Therefore, spatiotemporal models such as LSTM networks and RNNs are quite popular (Hu *et al.*, 2019). For example, an attention-based RNN model can achieve a clear and effective representation of time-space relationships and its learning ability is superior to that of other methods for both short- and long-term predictions of dissolved oxygen (Liu *et al.*, 2019). These models can be continuously optimized during the prediction process to improve their prediction accuracies (Deng *et al.*, 2019).

The prediction of dissolved oxygen and other water quality parameters is closely related to time. Attention-equipped, LSTM, DBN, and other DL models are able to mine the time sequence information well and achieve satisfactory results. Therefore, how to use DL models to avoid or reduce the negative impact of uncertainty factors on prediction results will be an important development direction in water quality prediction tasks.

Table 6 Water quality prediction

| | Field | Model | Framework | Data | Preprocessing augmentation | Transfer learning | Evaluation index | Results | Comparisons with other methods |
|---|---|---|---|---|---|---|---|---|---|
| 1 | Ta and Wei (2018) | CNN, LSTM | TensorFlow | 4,500 samples were collected from Mingbo Aquatic Products Co. Ltd. | NA | N | MSE | The accuracy and stability are sufficient to meet actual demands. | BP (traditional BP, MSE = 0.04, Holt-Winters α = 0.4, MSE = 0.06) |
| 2 | Liu *et al.* (2019) | RNN | PyTorch | A total of 5,006 sets were collected from a pond. | NA | N | RMSE, MAPE, MAE | The attention-based RNN can achieve more accurate prediction | SVR-linear, SVR-rbf, MLP, LSTM, Encoder-decoder, Input-Attn, DARNN, GeoMAN, Temporal-Attn, Spatiotemporal |
| 3 | Lin *et al.* (2018) | DBN | NA | 708 water samples were collected in twelve shrimp culture ponds. | NA | N | RMSE, WQI | Accuracy of model is satisfied | NA |
| 4 | Hu *et al.* (2019) | LSTM | Tensorflow | Data collection was achieved by deploying sensor devices in a cage. | Data filling and correction | N | Accuracy, time cost | prediction accuracy: pH: 95.76%; temperature: 96.88% | The proposed method can achieve a higher prediction accuracy and lower time cost than the RNN-based prediction model |
| 5 | Deng *et al.* (2019) | LSTM | NA | The data are three representative shrimp ponds | Data normalization | N | Accuracy | DopLstm achieves the highest accuracy | CF, AR, NN, SVM, and GM |
| 6 | Ren *et al.* (2020) | DBN | NA | Sensors were set up to collect data collect every 10 min with a result of 12,700 instances of data. | VMD algorithm | N | $R_2$ | 0.9336 | Bagging: 0.9014; Adaboost: 0.9262; Decision tree: 0.9189; CNN: 0.8811 |

## 4. Technical details and overall performance

The data and algorithms used are the two main elements of AI (Thrall *et al.*, 2018). These elements are all necessary conditions for AI to achieve success.

### 4.1 Data

In DL, an annotated dataset is critical to ensure a model's performance (Zhuang *et al.*, 2019). However, in practice, dataset construction is often affected by issues related to both quantity and quality. Before any images or specific features can be used as the input to a DL model, some effort is usually necessary to prepare the images through preprocessing and/or augmentation. The most common preprocessing procedure is to adjust the image size to meet the requirements of the DL model being applied (Sun *et al.*, 2018 ; Siddiqui *et al.*, 2017). In addition, the learning process can be facilitated by highlighting the regions of interest (Wang *et al.*, 2017 ; Zhao *et al.*, 2018b), or by performing background subtraction, foreground pixel extraction, image denoising enhancement (Qin *et al.*, 2016 ; Zhao *et al.*, 2018b ; Siddiqui *et al.*, 2017) and other steps to simplify image annotation.

Additionally, some related studies have applied data augmentation techniques to artificially increase the number of training samples. Data augmentation can be used to generate new labeled data from existing labeled data through rotation, translation, transposition, and other methods (Meng *et al.*, 2018 ; Xu & Cheng, 2017). These additional data can help to improve the overall learning process; and such data augmentation is particularly important for training DL models on datasets that contain only small numbers of images (Kamilaris & Prenafeta-Boldú, 2018).

In addition, to avoid being constrained by the limited availability of annotation data, some scholars have directly used pretrained DL models to conduct fish classification, thus avoiding the need to acquire a large volume of annotated data (Ahmad *et al.*, 2016). However, this approach has many limitations, such as negative transfer (Pan & Yang, 2010), learning or not learning from holistic images (Sun *et al.*, 2019), and is consequently difficult to implement satisfactorily for specific applications; hence, it is typically suitable only for theoretical algorithm research.

### 4.2 Algorithms

(1) Models. From a technical point of view, various CNN models are still the most popular (29 papers, 71%). However, 2 of the papers reviewed here use a GAN, 3 use an RNN, 2 use an LSTM, 2 use both an LSTM and a CNN, and 2 papers use a DBN and YOLO, respectively. Some CNN models are combined with output-layer classifiers, such as SVM and Softmax (Qin *et al.*, 2016； Sun *et al.*, 2018) or Softmax (Zhao *et al.*, 2018b； Naddaf-Sh *et al.*, 2018) classifiers.

(2) Frameworks. Caffe and TensorFlow are the most popular frameworks. One possible reason for the widespread use of Caffe is that it includes a pretrained model that is easy to fine-tune using transfer learning (Bahrampour *et al.*, 2015). Whether used for specific commercial applications or experimental research, the combination of DL and transfer learning helps to reduce the need for a large amount of data while saving significant training time (Erickson *et al.*, 2017). In addition, a variety of other DL frameworks and datasets exist that users can use easily. In particular, because of its strong support for graphical processing unit (GPU), the PyTorch framework has been used extensively in relatively recent literature (Ketkar, 2017； Liu *et al.*, 2019).

In fact, much of the research reviewed here (9/41) uses transfer learning (Siddiqui *et al.*, 2017； Levy *et al.*, 2018； Sun *et al.*, 2018), which involves using existing knowledge from related tasks or fields to improve model learning efficiency. The most common transfer learning technique is to use pretrained DL models that have been trained on related datasets with different categories. These models are then adapted to the specific challenges and datasets (Lu *et al.*, 2015). Figure 7 shows a typical example of transfer learning. First, the network is trained on the source task with the labeled dataset. Then, the trained parameters of the model are transferred to the target tasks (Sun *et al.*, 2018； Oquab *et al.*, 2014).

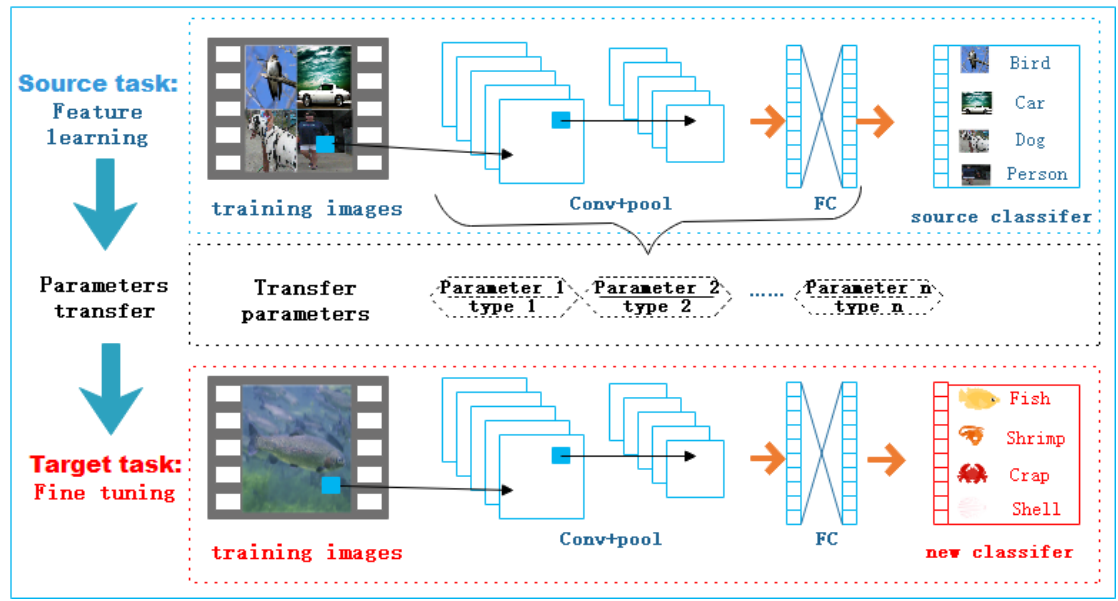

Figure 7. Typical example of transfer learning

(3) Model inputs. Although some studies use fish audio and water quality data, most of the model inputs are images (34, 83%). This situation reflects the significant advantage offered by DL in data processing, especially image processing. The inputs include public datasets such as the ImageNet dataset, the Fish4Knowledge (F4K) dataset, and the Croatian and Queensland University of Technology (QUT) fish datasets. Other datasets include data collected and produced in the field or obtained through Internet search engines, such as Google (Meng *et al.*, 2018 ; Naddaf-Sh *et al.*, 2018). Combining optical sensors and machine vision with DL systems provides possibilities for developing faster, cheaper and noninvasive methods for in situ monitoring and post-harvesting quality monitoring in aquaculture (Saberioon *et al.*, 2017). However, whether these datasets consist of text, audio, or image/video data, they typically hold large volumes of data. Such large amounts of data are particularly important when the problem to be solved is complex or when the difference between adjacent classes is small.

(4) Model outputs. Among the models used for classification, the outputs range from 4 to 16 classes. For example, one study considers images of 16 species of fish, and another considers 4 types of fish sound files. Among the other papers, 13 targeted live fish recognition where the outputs were fish and nonfish; 7 were size or biomass estimations; 2 were quantifications of fish feeding intensity; 6 were water quality predictions; and 5 were behavior analyses. However, from a technical point of view, the boundaries for identification, classification, and biomass estimation based on these classification models are quite vague. In these papers, the output and input classes for each model are the same. Each output consists of a set of probabilities that each input belongs to each class, and the model finally selects the class with the highest output probability for each input as the predicted class of that input.

## 4.3 Performance evaluation indexes and overall performance

### 4.3.1 Performance evaluation indexes

A variety of model performance evaluation indexes used in the literature are listed in Table 7. Most recognition and classification studies use common machine learning evaluation indicators such as accuracy and precision (Siddiqui *et al.*, 2017； Qin *et al.*, 2016). In behavior trajectory tracking, indicators such as the miss ratio (MR) are used (Wang *et al.*, 2017； Xu & Cheng, 2017). When water quality prediction is performed, additional indicators such as the mean absolute percentage error (MAPE) and root mean square error (RMSE) are used (Liu *et al.*, 2019). Moreover, a program's running speed is also an important performance indicator, especially when high real-time performance is required (Villon *et al.*, 2018； Zhou *et al.*, 2017a).

Because of the differences in the models, raw data, hardware operating environments, and parameters used in different studies, it is unscientific to compare different models based on only one parameter (Tripathi & Maktedar, 2019). However, in general, most of the studies in which the accuracy is used as a performance evaluation index report values above 90%, some even reach almost 100% (Banan *et al.*, 2020； Romero-Ferrero *et al.*, 2019), indicating that these method perform well. Among the papers using precision and recall as evaluation indexes, the highest results to date are 99.68% and 99.45%, respectively, which illustrates the advantages of DL models.

Table 7 Performance evaluation indexes for DL models

| Performance evaluation index | Description |
|---|---|
| Accuracy | Accuracy is the ratio of the number of correctly predicted fish to the total number of predicted samples. |
| Precision | The ratio of correctly identified fish to the ground truth. |
| Recall | The ratio of correctly identified fish to the total identified objects. |
| Speed | The running time of the algorithm. |
| Intersection-over-Union | IOU is the overlap rate between candidate area and ground truth area. The ideal scenario is complete overlap (i.e., the ratio is unity). |
| (IOU) | |
| False positive rate (FPR) | FPR is the proportion of negative instances divided into positive classes to all negative instances. |
| Mean Squared Error (MSE) | The mean squared error is the expected value of the square of the difference between the parameter estimate and the true value. |
| Mean Coefficient of Variation (MCV) | The ratio of the standard deviation to the mean. The MCV reflects the degree of dispersion of two sets of data. |
| Mostly Tracked Trajectories (MT) | Percentage of ground truth which are correctly tracked more than 80% in length. Larger values are better |
| Mostly Lost Trajectories (ML) | Percentage of ground truth instances correctly tracked at less than 20% of their length. Smaller values are better. |
| Fragments (Frag) | Percentage of trajectories correctly tracked at less than 80% but at more than 20% of their length. |
| ID Switch | Average total number of times that a resulting trajectory switches its matched ground truth identity with another trajectory, the smaller the better. |
| Miss ratio (MR) | Percentage of fish that are undetected in all frames. |
| Error ratio | Percentage of wrongly detected fish in all frames. |
| Root Mean Square Error (RMSE) | RMSE is the square root of MSE. |
| F1-measure | The harmonic mean of precision and recall. |

## 4.3.2 Performance comparisons with other approaches

An important aspect of this review is to consider comparisons between DL and other existing approaches. However, most DL methods are related to image analysis, 7 DL models have been proposed based on water quality and audio data. These studies show that DL can handle a variety of data types in smart fish farming rather than only images. In general, a DL model can be considered better than other compared models only with regard to the same dataset and the same task.

When performing fish identification tasks, CNN models show an accuracy 18.5% higher than that of SVM models (Qin *et al.*, 2016), a precision 41.13% higher than that of Gabor filters and other similar feature extraction methods, and a precision 19.54% higher than that of linear discriminant analysis (LDA) and manual extraction (Sun *et al.*, 2018). In addition, a CNN model has been shown to be superior to id.Tracker (Wang *et al.*, 2017). Compared with the accuracy achievable by human experts (89.3%), the accuracy of a CNN model was been shown to be superior (95.7%) (Villon *et al.*, 2018). When estimating the age of the fish population in Moen *et al.* (2018), a CNN also showed better performance than human experts. The achieved mean coefficient of variation (CV) was 8.89%, which is considerably lower than the reported mean CV of human readings. This may be due to the availability of datasets in these areas, as well as to the unique characteristics of fish and other background features.

Compared with a backpropagation neural network (BPNN), a CNN model was measured to be 6.25% more accurate in feed intensity classification. The model evaluation index of this CNN model also improved compared with those of traditional manual feature extraction methods, such as FIFFB and SIFFB (Zhou *et al.*, 2017b ; Zhou *et al.*, 2019). Furthermore, the results of water quality prediction indicate that LSTM and attention-based RNN models achieve higher accuracy than has been achieved with a BPNN model, Holt-Winters forecasting, or a support vector regression (SVR) model based on either a linear function kernel (SVR-linear) or a radial basis function kernel (SVR-RBF) (Ta & Wei, 2018 ; Liu *et al.*, 2019).

In addition, GAN models typically achieve better overall performances in fish recognition compared with a CNN (Zhao *et al.*, 2018b).

## 5. Discussion

### 5.1. Advantages of deep learning

The key advantage of DL in aquaculture is that DL models perform better than do the traditional methods. This may be because traditional machine learning algorithms require the manual feature extraction from images. Manually selecting features is a laborious, heuristic approach, and the effect is highly dependent on both luck and experience (Mohanty *et al.*, 2016). In contrast, a DL algorithm can automatically learn and extract the essential features from images in a sample dataset. Such algorithms offer high accuracy and strong stability for irregular target recognition in complex environments (Daoliang & Jianhua, 2018), and they can effectively learn mappings and correlations between a sample and objects from that sample. In addition, useful features can be learned automatically using a general-purpose learning procedure (LeCun *et al.*, 2015).

For example, in fish recognition, a DL model can effectively extract essential fish features. Such models have shown strong stability under challenging conditions such as low light and high noise, and they perform better than do traditional artificial feature extraction methods (Sun *et al.*, 2018). In behavioral analysis research, a DL model can effectively address problems related to occlusion (Wang *et al.*, 2017). In addition, a DL model can be used not only to monitor unknown objects or anomalies but also to predict parameters such as water quality.

Although DL models require more computing power and longer training times than do traditional methods (such as the SVM and random forest methods), after training is complete, the trained DL models are highly efficient at performing test tasks. For example, in a fish recognition study (Villon *et al.*, 2018), using for 900,000 images, the training process lasted 8 days on a computer with 64 GB of RAM, an i7 CPU @3.50 GHz, and a Titan X GPU card. However, after training was complete, the recognition time for each frame was only 0.06 s. In a study by Ahmad *et al.* (2016), training the CNN model required 5~6 h without a GPU implementation. However, during testing, each fish image required only approximately 1 ms for classification, making this model fully compatible with real-time processing needs.

## 5.2. Disadvantages and limitations of deep learning

At present, DL technology is still in a weak AI stage (Lu *et al.*, 2018). While weak AI systems can simulate the functions of the mind though a computationally system; however, they cannot yet artificially recreate a mind (Di Nucci & McHugh, 2006). The ability of DL models to constantly learn and improve is still very weak. In smart fish farming, DL models are used only as "black boxes". Because DL models are excessively reliant on sample data and have low interpretability, they can typically gain experience only from a specific dataset. Moreover, when faced with unbalanced training data, most models will tend to ignore some important features (Zhang & Zhu, 2018).

(1) Incomplete data. One of the most significant drawbacks of DL is the large amount of data required during training. For example, when using DL to identify fry size, not only are there many kinds of fish but their body shape and posture of each growth stage are also quite different, which necessitates high requirements for data collection and DL training. However, in traditional fisheries, no such datasets exist, or the available datasets are not sufficiently comprehensive. Thus, in this initial stage, much basic data collection work remains to be done.

Although data augmentation technology can be used to add some labeled samples to an existing dataset, when dealing with complex problems (e.g., multiclass problems with high precision requirements), more diversified training data are needed to improve accuracy (Patrício & Rieder, 2018). Because data annotation is a necessary operation in most cases, some complex tasks require experts to annotate data, and such expert volunteers are prone to make mistakes during data annotation, especially for challenging tasks such as fish species identification (Hanbury, 2008 ; Bhagat & Choudhary, 2018). Furthermore, data preprocessing is often a necessary and time-consuming task in DL, whether for image or text data (Choi *et al.*, 2018). In addition, some existing datasets do not fully represent the problems toward which they are oriented. Finally, in the field of smart fish farming, researchers do not have access to many publicly available datasets; thus, in many cases, they need to develop custom image sets, which can take hours or days of work.

(2) High cost. Whether the people involved are AI technicians or farmers, a large number of sensors need to be deployed when collecting data, making the up-capital cost investment in the early stage substantial. Another limitation is that DL models demand high levels of computing power; in fact the available common CPUs are typically unable to meet the requirements of DL (Shi *et al.*, 2016). Instead, GPUs and tensor processing units (TPUs) are the mainstream sources of computing power suitable for DL; consequently, the hardware requirements are very high, and the cost is also quite high (Wei & Brooks, 2019). In the absence of expected results, it is more difficult to persuade farmers to invest in the intelligent breeding industry, which is true in every country.

## 5.3. Future technical trends of deep learning in smart fish farming

(1) The applications of DL in aquaculture will continue to expand or emerge. Various existing applications of DL in smart fish farming are covered in this review. The current application fields include live fish identification, species classification, behavioral analysis, feeding decisions, size or biomass estimation, and water quality prediction. Other possible application areas with great potential include fish disease diagnosis, aquatic product quality safety control and traceability, although no relevant research has been reported to date. For example, tools that automatically diagnose fish diseases and provide reasonable suggestions for managing identified diseases are expected to be an important application area, especially in relation to image processing.

(2) Available dataset is becoming increasingly important. Datasets are an increasingly dominant concern in DL sometimes even more important than algorithms. With the improved transparency of aquaculture information and the establishment of open fishery databases, researchers will be able to access a broader variety of sample data more easily. Although the number of publicly available datasets is still small, Appendix A lists some datasets that are freely available for download. Researchers can use these datasets to test their DL models or to pretrain DL models and then adapt them to more specific future challenges. Due to the limited dataset availability and the difficulty of collecting real data, methods of improving the recognition rate from small numbers of samples represents an inevitable direction for future research. Transfer learning can be used to ameliorate the problem of insufficient sample data. Additionally, the necessary preprocessing and augmentation of datasets will become increasingly important.

(3) More advanced and complex models will continue to improve the performance of deep learning tasks. A combination model can be used to solve many of the problems faced by single models; as a result, more complex architectures will emerge. All types of DL models and classifiers as well as handcrafted features can be combined to improve the overall results. CNNs are widely used, but they consider each frame independently and ignore the time correlations between adjacent frames. Therefore, it is necessary to consider models that can account for spatiotemporal sequences. It is expected that in the future, more methods similar to LSTM networks or other RNN models will be adopted to achieve higher classification or prediction performances that capitalize on the time dimension. Examples of such applications include estimating fish growth based on previous continuous observations, assessing fish water demands, developing measures to avoid disease, and fish behavior analysis. Such models can also be applied in environmental studies to predict changes in water quality. Finally, some of the solutions discussed in this paper may become commercially available soon.

## 6. Conclusion

This paper conducted a deep and comprehensive investigation of the current applications of deep learning (DL) for smart fish farming. Based on a review of the recent literature, the current applications can be divided into six categories: live fish identification, species classification, behavioral analysis, feeding decisions, size or biomass estimation, and water quality prediction. The technical details of the reported methods were comprehensively analyzed in accordance with the key elements of artificial intelligence (AI): data and algorithms. Performance comparisons with traditional methods based on manually extracted features indicate that the greatest contribution of DL is its ability to automatically extract features. Moreover, DL can also output high-precision processing results. However, at present, DL technology is still in a weak AI stage and requires a large amount of labeled data for training. This requirement has become a bottleneck restricting further applications of DL in smart fish farming. Nevertheless, DL still offers breakthroughs for processing text, images, video, sound and other data, all of which can provide strong support for the implementation of smart fish farming. In the future, DL is also expected to expand into new application areas, such as fish disease diagnosis; data will become increasingly important; and composite models and models that consider spatiotemporal sequences will represent the main research direction. In brief, our purpose in writing this review was to provide researchers and practitioners with a better understanding of the current applications of DL in smart fish farming and to facilitate the application of DL technology to solve practical problems in aquaculture.

## Acknowledgments

The research was supported by the National Key Technology R&D Program of China (2019YFD0901004), the Youth Research Fund of Beijing Academy of Agricultural and Forestry Sciences (QNJJ202014), and the Beijing Excellent Talents Development Project (2017000057592G125).

Appendix A: Public dataset containing fish

| NO | Dataset | URL | Description | References |
|---|---|---|---|---|
| 1 | Fish4-Knowledge | http://groups.inf.ed.ac.uk/f4k/index.html | This underwater live fish dataset was acquired from a live video dataset captured in the open sea. It contains a total of 27,370 verified fish images in 23 clusters. Each cluster is represents a single species. | Boom *et al.* (2012) |
| 2 | Croatian fish dataset | http://www.inf-cv.uni-jena.de/fine_grained_recognition.html#datasets | This dataset contains 794 images of 12 different fish species collected in the Adriatic sea in Croatia. All the images show fishes in real-world situations recorded by high definition cameras. | Jäger *et al.* (2015) |
| 3 | LifeCLEF14 and LifeCLEF15 dataset | http://www.imageclef.org/ | The LCF-14 dataset for fish contains approximately 1,000 videos. Labels are provided for approximately 20,000 detected fish in the videos. A total of 10 different fish species are included in this dataset. LifeCLEF 2015 (LCF-15) was taken from Fish4Knowledge. LCF-15 consists of 93 underwater videos covering 15 species and provides 9,000 annotations with species labels. | Ahmad *et al.* (2016) |
| 4 | Fish-Pak | https://doi.org/10.17632/n3ydw29sbz.3#folder-6b024354-bae3-460aa758-352685ba0e38 | This is a dataset consisting of images of 6 different fish species i.e., *Catla* (Thala), *Hypophthalmichthys molitrix* (Silver carp), *Labeo rohita* (Rohu), *Cirrhinus mrigala* (Mori), *Cyprinus carpio* (Common carp) and *Ctenopharyngodon idella* (Grass carp). | Rauf *et al.* (2019) |
| 5 | ImageNet | http://www.image-net.org/ | ImageNet is an image database organized according to the WordNet hierarchy (currently only nouns), in which each node in the hierarchy is associated with hundreds or thousands of images. ImageNet currently has an average of over five hundred images per node. | Deng *et al.* (2009) |